\documentclass{article}

\usepackage{ijcai23}

\usepackage{times}
\usepackage{soul}
\usepackage{url}
\usepackage[hidelinks]{hyperref}
\usepackage[utf8]{inputenc}
\usepackage[small]{caption}
\usepackage{graphicx}
\usepackage{amsmath}
\usepackage{amsthm}
\usepackage{booktabs}
\usepackage{algorithm}
\usepackage{algorithmic}
\usepackage[switch]{lineno}
\usepackage{multicol}
\usepackage{multirow}
\usepackage{textcomp}
 \usepackage[table,xcdraw]{xcolor}
 \usepackage{amsfonts}

\nolinenumbers

\title{MCNS: Mining Causal Natural Structures 
Inside Time Series via A Novel Internal Causality Scheme}

\author{
Yuanhao Liu$^{1,2}$
\and
Dehui Du$^{1,2}$\and
Zihan Jiang$^{1,2}$\And
Anyan Huang$^{1,2}$\And
Yiyang Li$^{1,2}$
\affiliations
$^1$Software Engineering Institute\\
$^2$East China Normal University, Shanghai, China\\
\emails
dhdu@sei.ecnu.edu.cn
}

\begin{document}

\maketitle

\begin{abstract}
Causal inference permits us to discover covert relationships of various variables in time series.
However, in most existing works, the variables mentioned above are the dimensions. The causality between dimensions could be cursory, which hinders the comprehension of the internal relationship and the benefit of the causal graph to the neural networks (NNs).
In this paper, we find that causality exists not only \emph{outside} but also \emph{inside} the time series because it implies the succession of events in the real world. It inspires us to seek the relationship between internal subsequences.
However, the challenges are the hardship of discovering causality from subsequences and utilizing the causal natural structures to improve Neural Networks.
To address these challenges, we propose a novel framework called Mining Causal Natural Structure (\textsl{MCNS}), which is \emph{automatic} and \emph{domain-agnostic} and helps to find the causal natural structures inside time series via the internal causality scheme.
We evaluate the \textsl{MCNS} framework and integrate NN with \textsl{MCNS} on time series classification tasks. Experimental results illustrate that our impregnation, by refining attention, shape selection classification, and pruning datasets, drives NN, even the data itself preferable accuracy and interpretability.
Besides, \textsl{MCNS} provides an in-depth, solid summary of the time series and datasets.
\end{abstract}

\section{Introduction}
Time series data, such as medical electrocardiograms and financial data, have played an essential role in society. Furthermore, the possibility of making causal inferences \cite{mastakouri2020causal,li2022learning} in time series data greatly appeals to social and behavioural scientists and has been widely used in a plethora of applications. However, classical causal discovery \cite{granger1969investigating} approaches in time series usually treat the time series as a whole, and problematic to find causal relationships inside time series.

A rich body of research has been proposed to seek causal relations in structured multivariate time series data. For example, most works suggested leveraging the concept of \emph{Granger} causality \cite{huang2019causal,schamberg2019measuring,mastakouri2021necessary}, and some other works proposed to rely on the idea of \emph{Pearl} causality in i.i.d multivariate time series data \cite{gerhardus2020high,bica2020time}. But those works focused on relations between dimensions, and we find that the causal relationships need to be more profound and in-depth. There is causality not only outside (as a whole) but also inside the time series(in the subsequence). The causal natural structure inside the time series is crucial for causal inference.

\begin{figure}[t]   
	
	\centering
	
	\includegraphics[width=0.9\linewidth,scale=1.0]{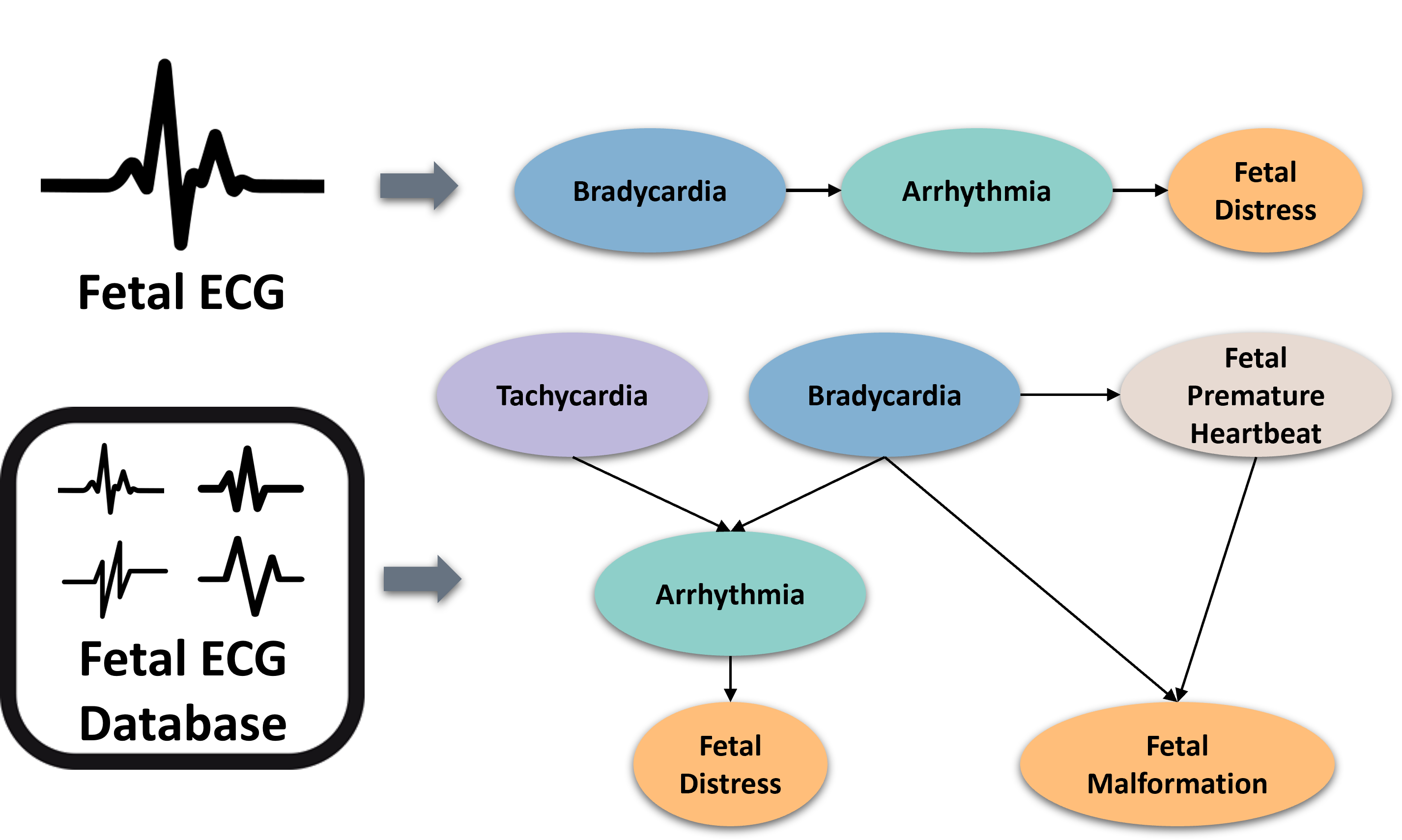}
	
	
	\caption{An example of a causal natural structure obtained from \textsl{MCNS} for the fetal ECG. The above is from a specific fetal ECG, and the below is from the whole dataset.}
	
	\label{intro}
	
\end{figure}

Actually, discovering causal relationships inside time series is also valuable or vital for making decisions.
For instance, when a medical AI system assists doctors in dealing with the classification of diseases in the fetal electrocardiogram (ECG), causal inference could help to figure out the exact distinguishable subsequences (symptoms) crucial for accurate and explainable diagnosis.
As shown in Figure 1, if the system can spot two crucial points, (1) the cause chain of the disease from a given specific fetal ECG, and (2) obtaining causal natural structures from the fetal ECG database, then the prediction disease can be more convincing and helpful, also straightforward to locate errors in the AI, rather than a label from a black box.
In practice, we expect a medical AI system to provide human-readable and sound explanations to support doctors in making the right decisions. It is worthwhile, especially for underdeveloped areas, where such techniques could help the doctors of rural areas with more reliable references from previous cases. Furthermore, our approach is domain agnostic, which would be applied to other domains, such as autonomous driving, the financial field, etc.
Therefore, an intuitive idea we want to explore is, can we discover causality, not from the relation between dimensions, but from inside specific time series of one dimension? 
However, there are two challenges: (1) How to discover causality inside time series? (2) If there is a causal relationship inside the time series, how to leverage it to benefit neural networks?

To deal with these issues, we propose a novel framework called Mining Causal Natural Structures (\textsl{MCNS}). We discover representative subsequences called snippets from the time series and utilize snippets to encode the initial time series into a binary sequence for discretizing a continuous time series. Then, we use a Greedy Fast Causal Inference (GFCI) algorithm to seek causal relations between snippets and construct an inside causal graph. It is worth mentioning that, unlike most related work that requires domain knowledge, \textsl{MCNS} is \emph{domain-agnostic}, which greatly enhanced the generalization of our approach. 

Based on the above explorations, we do not follow the existing causal graph construction approach that requires pruning or constructing by domain experts, which is a non-automated causal discovery. We impose two restrictions on the GFCI algorithm so that it can \emph{automatically} prune causal graphs. After that, we determine the final causal natural structure using the Bayesian Information Criterion (BIC) and calculate causal strength on edges using Propensity Score Matching (PSM) \cite{caliendo2008some} and Average Treatment Effect (ATE) \cite{holland1986statistics}.

For the second challenge, we impregnate Deep Neural Networks (DNNs) with causal graphs generated by \textsl{MCNS}. The first usage is inspired by \cite{jain2019attention,serrano2019attention}, which confirmed that attention could not correctly communicate the relative importance of inputs. Hence, we employ causal strength to refine attention to be more precise. Secondly, we leverage the \textsl{MCNS} to select shapes to classify time series, similar to the shapelets-based classification method but more explainable and accurate. Additionally, we prune the dataset with the portion containing causality, which leaves the most critical part and results in more accuracy and efficiency.

Our evaluation based on the PyTorch framework with the UCR dataset demonstrates that our \textsl{MCNS} can successfully inject the extracted causal knowledge into deep neural networks and improve NN's performance extensively, especially accuracy and interpretability. 

In summary, our main contributions are as follows:
\begin{itemize}
    \item We propose a novel framework for mining causal natural structures (\textsl{MCNS}) inside time series, which is both domain-agnostic and automatic. 
    \item We investigate training popular neural network models with our causal natural structures obtained from \textsl{MCNS}. It boosts neural network models to realize causal knowledge emanating from \textsl{MCNS}.
    \item Experimental results illustrate that our \textsl{MCNS} can effectively enhance NN models for better performance in time series of various domains and scales. It can also help improve the interpretability of neural networks and the time series itself. 
\end{itemize}

\section{Related Work}
\textbf{Causal Inference in Time Series.} There has been a propensity toward creating algorithms for causal inference on time series data.
A mainstream of works is based on domain knowledge, artificially constructing causal graphs to solve time series problems in a particular field \cite{deng2021compass,li2021causal,mathur2022exploiting}. Despite the success of these works in their respective fields, they involved bringing in domain experts to build causality relations rather than automating the discovery of causality in time series. 
Moreover, for mining the causality in the time series, some works use Granger causality to analyze the time series \cite{huang2019causal,schamberg2019measuring,mastakouri2021necessary}. Since these works use it to explore interactions inside time series, aside from the fact that it is actually investigating the causality between time series dimensions, Granger causality only means causality in the statistical sense, and it can not judge the internal mechanism between time series. To our knowledge, it is the first work on univariate time series causal discovery. We take causality between subsequences into account, in other words, mining causal natural structures inside time series, which is the main novelty of our work.

\textbf{Time Series Natural Structure.} Finding natural structure in time series is a significant issue. Some works attempted to solve this problem using probabilistic models, such as Autoplait \cite{matsubara2014autoplait} and TICC \cite{hallac2017toeplitz}. Some works also used change point detection \cite{Samaneh2019real,dette2020likelihood,xu2021optimum,lu2021grab}.
However, the existing work needs to illustrate profound relations between subsequences.
This paper utilizes causality to address the time series natural structure discovery problem. We propose a novel causal discovery framework to construct causal natural structures from univariate time series. 

\textbf{Causal Inference for Neural Networks.} Recently, researchers have attempted to study relations between causality and neural networks \cite{gao2019modeling,weber2020causal,liu2021everything,geiger2021causal,zevcevic2021interventional}. Since causality is widely used, various studies have applied causal structure as a part of NN or apply NN for causal discovery. However, most directly proposed a causal graph or utilize causal thinking. In this paper, we use inside causal graphs obtained from \textsl{MCNS} to enhance NN models in time series, which can boost their accuracy and interpretability.

\begin{figure*}[t]   
	
	\centering
	
	\includegraphics[width=\linewidth,scale=1.00]{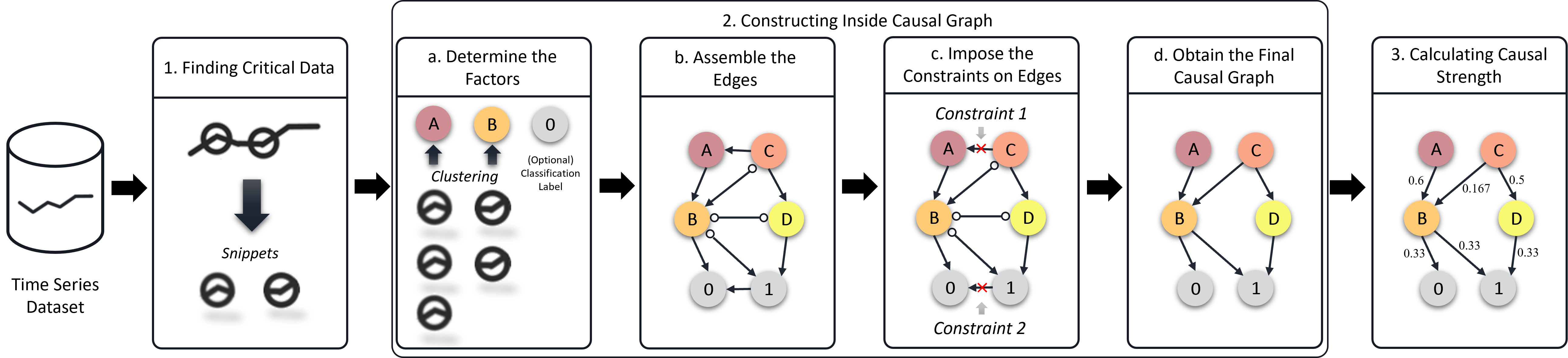}
	
	
	\caption{Overview of the \textsl{MCNS} approach. In step 1, time series snippets as representative data (thick black circles) are mined from the entire time series. In step 2a, snippets are clustered as different factors, and add an optional label at the end. In steps 2b and 2c, edges are assembled by the GFCI algorithm, and edges containing a hollow circle(e.g., $\circ\! \rightarrow$ and $\circ\!-\!\circ$) are determined as arrows without hollow circles (e.g., $\rightarrow$ and $none$). Edges are pruned by constraints. The final causal natural structure is pruned by the bayesian information criterion in step 2d. Causal strength is calculated in the step 3.}
		\label{framework}
	
\end{figure*}

\section{Approach}
Our mining causal natural structures framework has three components: finding critical data in time series, constructing inside causal graph, and calculating causal strength. Figure 2 shows the overall architecture of our approach.

\subsection{Problem Definition}\label{App0}
\textsl{MCNS} is used to find a causal natural structure $\mathbb{S}$ in subsequences $T_{i, m}$ from given time series $T$ as follows:
\begin{itemize}
    \item A time series $T$ is a sequence of real-valued numbers $t_i: T=t_1, t_2, \ldots, t_n$, (with an optional label $l_T$ for classification tasks), where $n$ is the length of $T$. 

    \item A subsequence $T_{i, m}$ of a time series $T$ is a continuous subset of the values from $T$ of length $m$ starting from position $i$. Formally, $T_{i, m}=t_i, t_{i+1}, \ldots, t_{i+m-1}$, where $1 \leq i \leq n-m+1$.
    
    \item A causal natural structure $\mathbb{S}$ inside time series $T$ is a 4-tuple $<S_{sub}, l_T, \psi, C>$, which is composed of subsequences set $S_{sub}$, optional label $l_T$, causal relations $\psi$, and causal strength $C$.
\end{itemize}

\subsection{Finding Critical Data in Time Series}\label{App1}
 First, we should find critical data representing the entire time series to discover the event in the real world behind it.

To begin with, we need to determine how to set the subsequence length $l$. Since the subsequence length corresponds to the time span of events occurring, and often similar real-world events have periodicity, it is desirable that $l$ is equal to the length of the intrinsic period of the time series $T$. For example, concerning fetal ECG data shown in Figure 1, $l$ should be around the duration of a single fetal heartbeat. We adopt the popular Fast Fourier Transform (FFT) \cite{cooley1965algorithm} as a solution. Time series $T$ is converted into the frequency domain, extracting the dominant frequency $f$. The subsequence length $l$ is guided by $1/f$. In section 5, we will explore the effectiveness of our approach.

Additionally, to determine the same subsequence length of the complete dataset, we calculate the subsequence length $l_T$ for each time series $T$ in the dataset using FFT, and employ the maximum value in $l_T$ as the unified subsequence length.

To extract representative subsequences, we discover $k$ snippets $s_{T}$ from each time series $T$ using the time series snippets algorithm \cite{imani2018matrix}, which is domain agnostic to guarantee \textsl{MCNS} can be applied to datasets in any domain.

\subsection{Constructing Inside Causal Graph}\label{App2}
In order to construct the causal graph, we should determine the factors, assemble the edges between factors, impose the constraints on edges, and obtain the final causal graph.

\subsubsection{Determine the Factors}
To merge similar subsequences, we cluster the subsequences obtained in the previous step into $n$ classes using $k$-shape clusters algorithm \cite{paparrizos2015k}. These classes represent events mentioned above, as reflected by this dataset. $n$ classes factors and (optional) $l_T$ labels constitute the factors of the causal graph. Each time series $T$ can be expressed as a binary sequence. If $T$ contains the corresponding factors, the value of this factor is 1, and not vice versa \cite{liu2021everything}. This binary sequence represents the time series by events (e.g., Bradycardia, Arrhythmia, Fetal Distress in Figure 1).

\subsubsection{Assemble the Edges Between Factors} 
The following is to establish the edges, which denote the causal relationship between factors. We choose the GFCI algorithm \cite{ogarrio2016hybrid} to detect causal relations and infer without causal sufficiency. GFCI permits us to make causal inferences when having confounding variables and output a Partial Ancestral Graph (PAG). It offers us a preliminary inside causality between factors. 

\subsubsection{Impose the Constraints on Edges} 
Additionally, we put two kinds of constraints on the graph to refine the causal graph. The first is banning edge from label factors to other factors because classification labels are not involved in actual events. Moreover, the second is an effect that does not precede its cause \cite{black1956cannot}. What happens earlier in the time series leads to what happens later. For most time series, provided that factor $X$ appears after $Y$, we should ban the edge from $X$ to $Y$.

\subsubsection{Obtain the Final Causal Graph} 
The PAG obtained in the last part contains four edge types \cite{zhang2008causal}, which are $\rightarrow,\leftrightarrow,\circ\!\rightarrow,\circ\!-\!\circ$. Among them, $X \rightarrow Y$ denotes $X$ causes $Y$, and $X \leftrightarrow Y$ denotes that there is an unobserved confounder of $X$ and $Y$. So $\rightarrow$ edges are retained and $\leftrightarrow$ edges are removed. For the remaining two cases, $X \circ\!\!\rightarrow Y$ denotes either $X \rightarrow Y$ or $X \leftrightarrow Y$, and $X \circ\!-\!\circ Y$ denotes either $X \rightarrow Y$ , $Y \rightarrow X$ or $X \leftrightarrow Y$. There's no way to get a true probability, so we operate bootstrapping algorithm to determine the final causal graph.
Each case is given the same probability for the two uncertainties mentioned above. We employ Bayesian Information Criterion (BIC) \cite{schwarz1978estimating} to estimate the quality of each graph $G_n$ is measured by its fitness with time series $T$.

\subsection{Calculating Causal Strength}\label{App3}
Even after we have gone through the above steps, the resulting inside causal graph is still noisy. We calculate causal strength on edges to further refine the causal graph. High strength is allocated to edges with a high causal effect. Similarly, meager strength is allocated to edges with low causal effects.

We utilize propensity score matching to measure average treatment effect $\phi_{T,Y}$, which denotes the causal strength of $T \rightarrow Y$. In this paper, it represents the effect of changing the $1$ in our binary sequence to $0$ through the do-calculus on the classification result, that is to say, the effect of subsequence on classification: 
\begin{align}\label{eqn-9}
      \phi_{T, Y} &= E[Y \mid d o(T=1)]-E[Y \mid d o(T=0)]\\ 
      &= \left[\sum_{t_i=1} \Delta_{i, j} - \sum_{t_i=0} \Delta_{i, j}\right] / N
\end{align}

where the do-calculus $do(T = 1)$ shows intervention on $T$ and altering the value of $T$ to $1$. $j$ represents the most similar instance in a different set than $i$, and $\Delta_{i,j}$ means the difference between the outcome value of instance $i$ and $j$.

\subsection{Impregnation DNN with \textsl{MCNS}}\label{App4}
Recently, some effort has been made to exploit the DNNs, especially recurrent neural networks (RNNs), for different sizes of time series prediction and classification. However, the application of causal graphs to benefit deep neural networks in time series has been limited. We proposed three methods to impregnate DNN with \textsl{MCNS} as shown in Figure 3.

\begin{figure}[ht]   
	
	\centering
	
	\includegraphics[width=\linewidth,scale=1.00]{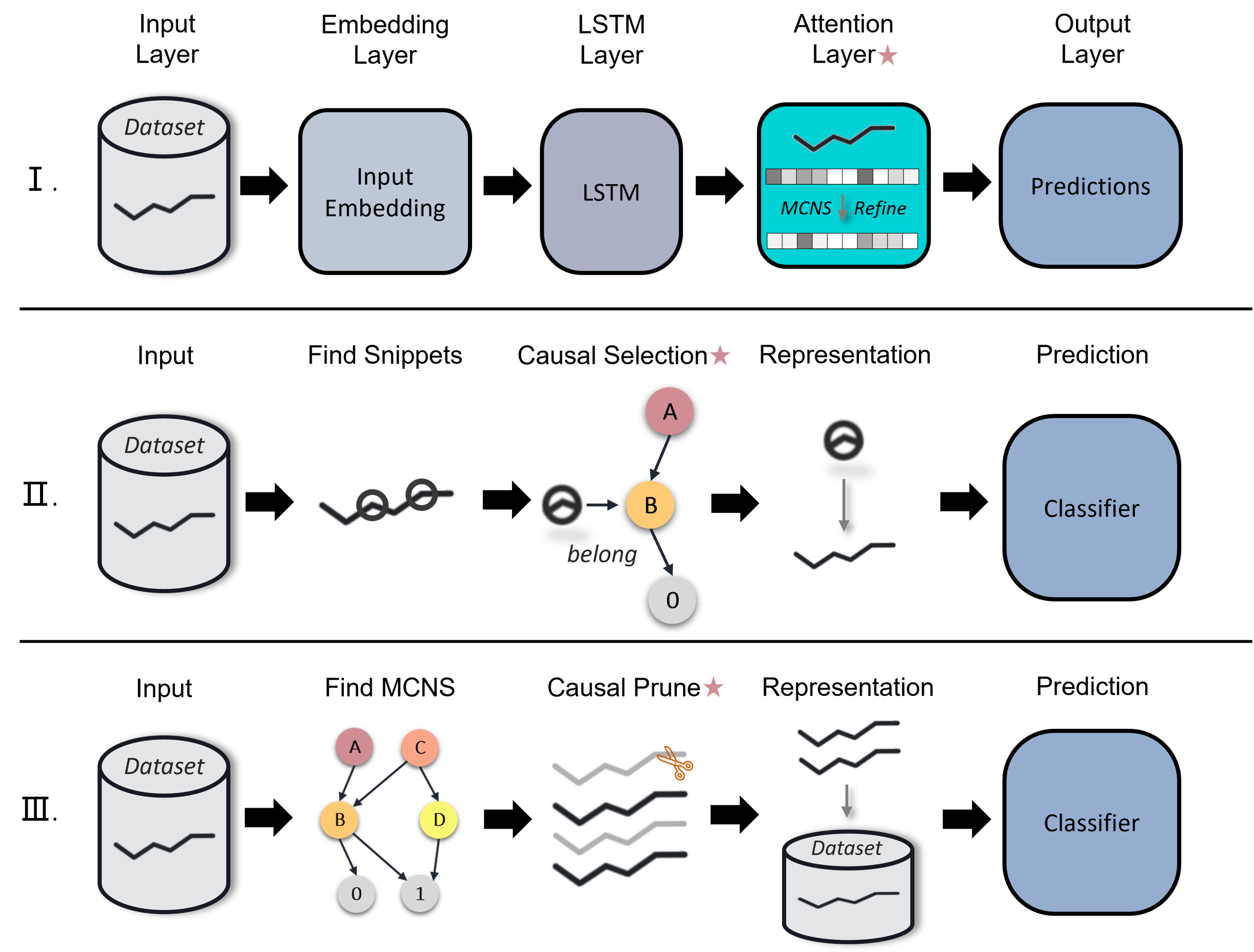}
	
	
	\caption{Three usages of impregnating neural networks with causal inference. \emph{Refine Attention with Causal Strength} (Above \uppercase\expandafter{\romannumeral1}), \emph{Shape Causal Selection Based Classification} (Middle \uppercase\expandafter{\romannumeral2}), and \emph{Dataset Prune with Causality}(Below \uppercase\expandafter{\romannumeral3}). The step with an asterisk is the core step of impregnation DNN with \textsl{MCNS}.}
	
	\label{intro}
	
\end{figure}

\subsubsection{Refine Attention with Causal Strength}
Attention has become an effective mechanism for superior results, as demonstrated in time series prediction and classification. However, some prior work substantiates that there is some distance away from attention and the relative importance of inputs \cite{jain2019attention,serrano2019attention}. Attention can not wholly explain the relative importance of inputs.

Causal strength can be exploited to improve it. We utilize a Long Short-Term Memory (LSTM) with attention model \cite{wang2016attention}. The input vector $X_{t-n} \cdots X_{t-3}, X_{t-2}, X_{t-1}, X_t$ is the $n$ multi-dimensional feature vectors up to the time to be predicted. The hidden layer processes the input vector in the LSTM in some intermediate states. The attention coefficient is obtained from the hidden layer of the last moment of another LSTM network in the decoder. Finally, vector $C$ passes the fully connected layer to calculate the predicted result vector $\varrho$.
\begin{equation}
    e_{i j}=\nu \tanh \left(W \cdot h_j+U \cdot h_{i-1}^{\prime}+b\right)
\end{equation}
\begin{equation}
    a_{i j}=\frac{\exp \left(e_{i j}\right)}{\sum_{k=t-n}^t \exp \left(e_{i k}\right)}
\end{equation}
\begin{equation}
    C=\sum_{j=t-n}^t a_{i j} h_j
\end{equation}

where $e_{i j}$ is the relation score between $h_{i-1}^{\prime}$ and $h_j$. $a_{i j}$ is the attention coefficient corresponding to $e_{i j}$. After that, the obtained attention coefficient is assigned to different middle layer states $h_j$ and summed to obtain vector $C$ input to the decoder. 

We refine attention using additional loss function $H_{cau}$. The extra loss function guides attention to the causal strength from \textsl{MCNS}. Specifically,
$\sum_{q=1}^Q \operatorname{BIC}\left(G_q, \mathbf{X}\right) \times \phi_{T, Y}$ is the causal strength corresponding to each factors, and $\zeta_i$ is the normalized strength over the whole time series. For the initial LSTM model, $H(p,q)$ means cross-entropy loss on $\varrho$. Hence, $H_{cau}$ and $H(p,q)$ can be denoted as follows:
\begin{equation}
    H(p, q)=-\sum_{i=1}^n p\left(x_i\right) \log \left(q\left(x_i\right)\right)
\end{equation}
\begin{equation}
    H_{cau} = \sum_{i=1}^n\lvert a_ {ij}-\zeta_i \rvert
\end{equation}

To sum up, we set the updated loss function of the model as follows:
\begin{equation}
    L = \alpha H(p, q) +\beta H_{cau} 
\end{equation}

where $\alpha + \beta = 1$. What is worth mentioning is that each factor represents a subsequence. The strength of all the time steps in this subsequence is treated as the causal strength of the factors.

 \begin{table*}[!t]
    \centering
    \begin{tabular}{lllllllll}
        \hline
        \textbf{Dataset}  & \textbf{Type} & \textbf{Abbr}  & \textbf{Brief Description} & \textbf{Train} & \textbf{Test} & \textbf{Prune} & \textbf{Class} \\
        \hline
        PowerCons & Power & PC & Electric power consumption & 180 & 180 & 171 & 2 \\
ECGFiveDays & ECG & EFD & ECG in two days & 23 & 861 & 14 & 2 \\
FordA & Sensor & FA & A car certain symptom(without noise) & 3601 & 1320 & 2056 & 2 \\
FordB & Sensor & FB & A car certain symptom(with noise) & 3636 & 810 & 2190 & 2 \\
Strawberry & Spectro & Sb & Food spectrographs & 613 & 370 & 442 & 2 \\
SmallKitchenAppliances & Device & SKA & Behavioural data   & 375 & 375 & 368 & 3 \\
        \hline
    \end{tabular}
    \caption{A summary of the benchmark datasets. The \emph{Prune} column states the size of the pruned train set for the CausalPrune method.}
    \label{tab:plain}
\end{table*}

\subsubsection{Shape Causal Selection Based Classification}
Causal inference explores how changes in variable $X$ affect another variable $Y$. When we set variable $X$ as the shapes inside the time series and variable $Y$ as the classification label, we can recognize that shapes affect the classification results. Hence, our causal natural structure from \textsl{MCNS} depicts the classification process of time series. 

In other words, \textsl{MCNS} may contain crucial information for time series classification. Hence, we operate the factors and causal relations in \textsl{MCNS} to guide the classification process in the neural network. Inspired by \cite{hills2014classification}, we leverage causal relations and snippets \cite{imani2018matrix} to classify time series.

For input time series $T$, we discover $k$ snippets $s_{iT}$, and concat snippets as a representation $r_T$ of a time series like shapelets:
\begin{equation}
    r_T=\operatorname{concat}\left(s_{1 T}, s_{2 T}, \ldots, s_{k T}\right)
\end{equation}

Furthermore, we draw on the causal graph to select snippets. If any snippets of time series belong to causal graph factors that affect the label, we choose them to represent the real content related to classification. If not, we utilize the initial $r_T$:
\begin{equation}
    r_T= \begin{cases}\operatorname{concat}\left(s_{j T}\right) & \text { if } s_{j T} \in S_{sub}, j \geq 1 \\ r_T & \text { otherwise }\end{cases}
\end{equation}

We mask the parts less than the maximum length and use them as input to the LSTM or other classifiers like $k$ nearest neighbor as the experimental setting. Hence, the neural network or traditional classifier can comprehend the nature of the input. 

\subsubsection{Dataset Pruning with Causality}
Every time series in the dataset only sometimes help NN to learn their features. Some data may be redundant or harmful \cite{angelova2005pruning}. However, causality can reveal the time series that matters for classification. We employ causality to prune the dataset.

To prune a time series dataset $\varsigma = \{ T_1,T_2,\ldots,T_m\}$, we first discover \textsl{MCNS} $S_\varsigma = <S_{sub},l_T,\psi,C>$ on the whole dataset and $S_{T_i} = <S_{i},l_{T_i},\psi_i,C_i>$ on each time series $T_i$.

Each time series $T_i$ in the dataset is treated as follows:
\begin{equation}\label{prune}
    T_i= \begin{cases}T_i & \text { if } <a_{ij},l_T> \in \psi, a_{ij} \in S_i, i \leq m  \\ None & \text { otherwise }\end{cases}
\end{equation} 

The equation (\ref{prune}) means that some time series are abandoned because their causal factors do not affect the classification label. The dataset is already pruned by the operations above.

Afterward, we got the essence data and input it to the LSTM or other neural networks. Therefore, the neural network can comprehend the entire dataset precisely through the essence data.  

\section{Experiment}

\subsection{Datasets}\label{Exp1}
 To illustrate that \textsl{MCNS} can be applied to datasets of different scales and multiple domains, we explore several experiments on six benchmark datasets introduced in the UCR time-series \cite{dau2019ucr}, which come from the electric power, biology, behavior, food spectrograph, and automotive subsystem domains. The details of the datasets are given in Table 1.

\begin{table*}[!h]
\scriptsize
\centering
\resizebox{\linewidth}{!}{%
\renewcommand\arraystretch{1.03}
\setlength{\tabcolsep}{1mm}{
\begin{tabular}{c|c|c|c|c|c|c|c} 
\toprule
 \multicolumn{1}{c|}{\textbf{Models}}	& \textbf{Ratio}	 & \textbf{PC}	& \textbf{EFD}	& \textbf{FA}	& \textbf{FB} & \textbf{Sb} &	\textbf{SKA}\\\midrule
\multirow{8}{*}{LSTM}&1$\%$&	61.67$\%$/55.06$\%$	&$-$	&49.16$\%$/43.77$\%$	&50.32$\%$/42.89$\%$&	57.30$\%$/67.08$\%$	&32.80$\%$/16.46$\%$\\
&5$\%$&	73.67$\%$/62.84$\%$	& $-$	&52.27$\%$/46.18$\%$	&52.47$\%$/37.93$\%$&	62.97$\%$/68.65$\%$	&33.06$\%$/16.56$\%$\\
&10$\%$&	85.56$\%$/85.41$\%$	&49.36$\%$/33.04$\%$	&55.83$\%$/55.04$\%$	&57.03$\%$/59.72$\%$&	57.03$\%$/51.96$\%$	&49.60$\%$/40.83$\%$\\
&30$\%$&	82.78$\%$/82.58$\%$	&51.45$\%$/37.30$\%$	&51.59$\%$/34.03$\%$	&62.96$\%$/60.64$\%$&	67.03$\%$/78.89$\%$	&38.40$\%$/28.22$\%$\\
&50$\%$&	91.11$\%$/91.09$\%$	&60.16$\%$/52.68$\%$	&59.69$\%$/66.16$\%$	&59.38$\%$/58.93$\%$&	68.38$\%$/76.17$\%$	&53.06$\%$/51.50$\%$\\
&80$\%$&	83.33$\%$/83.08$\%$	&80.60$\%$/80.16$\%$	&64.24$\%$/66.71$\%$	&61.39$\%$/61.72$\%$&	64.32$\%$/78.29$\%$	&53.60$\%$/51.84$\%$\\
&100$\%$&	92.77$\%$/92.77$\%$	&82.81$\%$/82.75$\%$	&60.23$\%$/64.21$\%$	&65.80$\%$/58.41$\%$&	72.97$\%$/77.97$\%$	&57.33$\%$/50.62$\%$\\
& Prune &	\textbf{97.44$\%$}/\textbf{97.44$\%$}	& \textbf{88.39$\%$}/\textbf{88.22$\%$}	&\textbf{86.06$\%$}/\textbf{86.04$\%$}	&\textbf{81.36$\%$}/\textbf{82.26$\%$}&	\textbf{87.56$\%$}/\textbf{90.25$\%$}	&\textbf{74.93$\%$}/\textbf{75.04$\%$}\\\midrule
\multirow{7}{*}{LSTM+Att}&1$\%$&	50.00$\%$/33.33$\%$	& $-$	&49.77$\%$/35.96$\%$	&53.82$\%$/51.04$\%$&	61.89$\%$/74.03$\%$	&33.06$\%$/16.59$\%$\\
&5$\%$&	50.00$\%$/33.33$\%$	& $-$	&51.59$\%$/34.03$\%$	&56.67$\%$/52.21$\%$&	64.32$\%$/39.14$\%$	&50.40$\%$/40.24$\%$\\
&10$\%$&	88.33$\%$/88.32$\%$	&49.71$\%$/33.20$\%$	&68.26$\%$/67.75$\%$	&57.65$\%$/61.59$\%$&	70.54$\%$/70.13$\%$	&60.53$\%$/60.27$\%$\\
&30$\%$&	87.22$\%$/87.16$\%$	&57.72$\%$/57.40$\%$	&\textbf{68.71$\%$}/\textbf{68.68$\%$}	&56.79$\%$/61.87$\%$&	65.94$\%$/65.93$\%$	&62.93$\%$/58.79$\%$\\
&50$\%$&	92.22$\%$/92.21$\%$	&74.09$\%$/72.99$\%$	&67.12$\%$/72.36$\%$	&68.20$\%$/63.09$\%$&	75.67$\%$/75.32$\%$	&66.13$\%$/65.34$\%$\\
&80$\%$&	93.89$\%$/93.88$\%$	&78.04$\%$/77.94$\%$	&71.21$\%$/74.34$\%$	&69.75$\%$/61.45$\%$&	76.75$\%$/74.92$\%$	&40.80$\%$/29.29$\%$\\
&100$\%$&	91.11$\%$/91.11$\%$	&84.32$\%$/84.08$\%$	&71.74$\%$/75.51$\%$	&69.50$\%$/65.69$\%$&	76.48$\%$/76.27$\%$	&39.73$\%$/33.55$\%$\\\midrule
\multirow{7}{*}{}&1$\%$&	68.33$\%$/66.42$\%$	& $-$	&49.84$\%$/48.01$\%$	&\textbf{54.32$\%$}/58.52$\%$&	\textbf{64.32$\%$}/39.14$\%$	&43.20$\%$/34.86$\%$\\
&5$\%$&	\textbf{90.00$\%$/89.99$\%$}	& $-$	&54.84$\%$/52.49$\%$	&\textbf{59.75$\%$}/54.17$\%$&	\textbf{64.86$\%$}/41.44$\%$	&\textbf{60.00$\%$}/48.00$\%$\\
&10$\%$&	\textbf{91.67$\%$}/\textbf{91.66$\%$}	&\textbf{63.47$\%$}/\textbf{60.30$\%$}	&\textbf{81.97$\%$}/\textbf{81.48$\%$}	&\textbf{58.47$\%$}/63.24$\%$&	\textbf{77.56$\%$}/76.85$\%$	&\textbf{64.00$\%$}/\textbf{63.41$\%$}\\
LSTM+Att&30$\%$&	\textbf{91.67$\%$}/\textbf{91.66$\%$}	&\textbf{68.29$\%$}/\textbf{64.95$\%$}	&65.34$\%$/61.28$\%$	&\textbf{66.41$\%$}/\textbf{69.91$\%$}&	\textbf{76.48$\%$}/75.59$\%$	&\textbf{65.06$\%$}/\textbf{62.62$\%$}\\
+MCNS&50$\%$&	\textbf{93.89$\%$}/\textbf{93.88$\%$}	&\textbf{75.49$\%$}/\textbf{75.45$\%$}	&\textbf{69.84$\%$}/\textbf{68.72$\%$}	&\textbf{68.89$\%$}/\textbf{70.91$\%$}&	\textbf{80.81$\%$}/79.26$\%$	&\textbf{67.73$\%$}/65.76$\%$\\
&80$\%$&	\textbf{94.44$\%$}/\textbf{94.44$\%$}	&\textbf{83.51$\%$}/\textbf{83.48$\%$}	&\textbf{74.62$\%$}/\textbf{74.94$\%$}	&\textbf{69.74$\%$}/\textbf{69.87$\%$}&	\textbf{82.43$\%$}/80.55$\%$	&63.20$\%$/56.26$\%$\\
&100$\%$&	\textbf{96.67$\%$}/\textbf{96.66$\%$}	&\textbf{89.54$\%$}/\textbf{89.34$\%$}	&\textbf{75.00$\%$}/\textbf{76.16$\%$}	&\textbf{72.67$\%$}/\textbf{73.19$\%$}&	\textbf{87.02$\%$}/\textbf{85.76$\%$}	&57.33$\%$/47.56$\%$\\\midrule
\multirow{7}{*}{MCNS}&1$\%$&	\textbf{73.13$\%$}/\textbf{69.32$\%$}	&$-$	&\textbf{55.30$\%$}/\textbf{52.34$\%$}	&52.22$\%$/61.03$\%$&	60.00$\%$/\textbf{70.98$\%$}	&\textbf{43.78$\%$}/\textbf{38.62$\%$}\\
&5$\%$&	75.56$\%$/76.59$\%$	&$-$			&\textbf{57.65$\%$}/\textbf{60.38$\%$}	&52.29$\%$/42.61$\%$&	62.60$\%$/65.35$\%$	&53.85$\%$/\textbf{49.43$\%$}\\
&10$\%$&	74.56$\%$/71.74$\%$	&50.29$\%$/66.92$\%$	&57.72$\%$/58.17$\%$	&56.17$\%$/55.79$\%$&	67.30$\%$/74.20$\%$	&58.67$\%$/55.49$\%$\\
&30$\%$&	75.56$\%$/69.86$\%$	&53.42$\%$/65.93$\%$	&57.12$\%$/61.34$\%$	&55.67$\%$/52.19$\%$&	74.59$\%$/\textbf{79.20$\%$}	&62.67$\%$/61.50$\%$\\
&50$\%$&	73.88$\%$/68.45$\%$	&60.28$\%$/63.76$\%$	&58.03$\%$/61.47$\%$	&57.63$\%$/53.67$\%$&	77.29$\%$/\textbf{81.25$\%$}	&66.67$\%$/\textbf{66.14$\%$}\\
&80$\%$&	76.67$\%$/73.42$\%$	&64.23$\%$/69.86$\%$	&60.23$\%$/62.34$\%$	&59.50$\%$/52.26$\%$&	79.72$\%$/\textbf{83.59$\%$}	&\textbf{70.40$\%$}/\textbf{66.93$\%$}\\
&100$\%$&	78.89$\%$/76.25$\%$	&64.69$\%$/64.57$\%$	&60.76$\%$/64.28$\%$	&58.89$\%$/56.36$\%$&	81.89$\%$/85.65$\%$	&\textbf{69.86$\%$}/\textbf{69.38$\%$}\\\midrule
\multirow{7}{*}{Shapelets}&1$\%$&	39.68$\%$/45.56$\%$	&$-$	&48.41$\%$/00.00$\%$	&49.50$\%$/\textbf{66.23$\%$}&	35.67$\%$/00.00$\%$	&32.78$\%$/11.47$\%$\\
&5$\%$&	43.89$\%$/56.28$\%$	&$-$			&48.41$\%$/00.00$\%$	&49.50$\%$/\textbf{66.23$\%$}&	64.32$\%$/\textbf{78.29$\%$}	&36.80$\%$/31.65$\%$\\
&10$\%$&	49.44$\%$/48.59$\%$	&48.89$\%$/55.73$\%$	&48.41$\%$/00.00$\%$	&49.50$\%$/\textbf{66.23$\%$}&	64.32$\%$/\textbf{78.29$\%$}	&33.87$\%$/17.78$\%$\\
&30$\%$&	52.22$\%$/51.51$\%$	&59.95$\%$/60.48$\%$	&51.59$\%$/68.07$\%$	&49.50$\%$/66.23$\%$&	64.32$\%$/78.29$\%$	&38.40$\%$/27.84$\%$\\
&50$\%$&	51.11$\%$/54.34$\%$	&49.71$\%$/00.00$\%$	&51.59$\%$/68.07$\%$	&49.50$\%$/66.23$\%$&	64.32$\%$/78.29$\%$	&34.13$\%$/18.36$\%$\\
&80$\%$&	50.00$\%$/66.67$\%$	&49.71$\%$/00.00$\%$	&51.59$\%$/68.07$\%$	&49.50$\%$/66.23$\%$&	64.32$\%$/78.29$\%$	&44.00$\%$/34.98$\%$\\
&100$\%$&	63.33$\%$/59.76$\%$	&49.71$\%$/00.00$\%$	&51.59$\%$/68.07$\%$	&49.50$\%$/66.23$\%$&	64.32$\%$/78.29$\%$	&42.67$\%$/42.67$\%$ \\
\bottomrule
\end{tabular}}}
    \caption{Performance on time series classification task. The first number is Acc, and the second number is F1. The highest results under each ratio are marked with bold. The \emph{Prune} line shows the performance of LSTM using the CausalPrune train set. The $-$ result indicates that the number of samples under this ratio is too small to reach the number of categories in the classification.}
\label{table:1}
\end{table*}

\subsection{Experimental Setup}\label{Exp1}

\textbf{Our Models.} In this paper, we evaluate our \textsl{MCNS} framework as described in Section 3.2-3.4 and three models impregnate NN with \textsl{MCNS} (\textsl{LSTM+Att+MCNS}, \textsl{MCNS}, and \textsl{CausalPrune}) as described in Section 3.5. 

\noindent\textbf{Parameter Settings.} We employ LSTM as the main body, and 2 hidden layers, 128 neurons in each layer, and one fully connected layer connected to the output function are uniformly set. Moreover, we find 5 snippets for each time series and the length determined by the FFT-based method.

\noindent\textbf{Comparison Models.} Because no previous work has found causality in univariate time series, we compare three \textsl{MCNS}-based models with NN baselines and shape-based methods, including \textsl{LSTM}, \textsl{LSTM+Att}, and \textsl{Shapelets} \cite{abelson-et-al:scheme}. \textsl{LSTM+Att} is a standard model of processing time series. Since the prior knowledge may result in unfair comparison, we do not add expert knowledge to keep our \textsl{MCNS} without domain experts involvement. 

\noindent\textbf{Other Settings.} The experiments are conducted on Windows 10, coming with an Intel Xeon Silver 4210R CPU and a NVIDIA Tesla T4 GPU.

\subsection{Main Results}\label{Exp4}
In this section, we investigate the classification accuracy and f1-score of our applications. Each set of experiments was repeated five times for \textsl{MCNS}, which is randomized. The main experimental results are shown in Table 2.

\textbf{Attention vs. No Attention.} We can see that \textsl{LSTM+Att} outperforms \textsl{LSTM} by around 3-4\% on average Acc and F1. However, sometimes the addition of attention will make the model less effective.  
That is because attention can only sometimes enhance the features that affect the results.
This above suggests that attention is helpful for neural network models to capture part of crucial information in the time series, but sometimes something else is needed.

\textbf{Attention vs. Attention + \textsl{MCNS}.} Furthermore, we can find out that
\textsl{LSTM+Att+MCNS} transcends \textsl{LSTM+Att} by around 6-8\% on average Acc and F1. The performance gap is related to the size of the dataset. The above illustrates that causal strength is helpful for attention-based models to discover core content in the time series. What is not explained in the table is that we find that \textsl{LSTM+Att+MCNS} converges much faster than \textsl{LSTM+Att}, which may be because attention has received the correct guidance.

\textbf{Causal Inference vs. Neural Networks.} Comparing \textsl{MCNS} with NN baselines \textsl{LSTM} and \textsl{LSTM+Att}, we observe in few-shot settings $(1\%, 5\%)$, \textsl{MCNS} outperforms NNs by about 6-7\% on average Acc and 12-18\% on average F1 since NNs tend to underfit in few-shot settings. However, with the increase in training data, the performance gap becomes narrower, and consequently, NNs outperform \textsl{MCNS} in several cases.
Compared with \textsl{MCNS}, NNs have the advantage of learning from large amounts of data.

\textbf{\textsl{MCNS} vs. \textsl{Shapelets}.} Similarly, \textsl{MCNS} and \textsl{Shapelets} are both discriminative subsequences for time series classification. Comparing \textsl{MCNS} with baselines \textsl{Shapelets} in the case where the other settings are the same except for the shape selection, we observe that \textsl{MCNS} outperforms \textsl{Shapelets} by about 14\% on average Acc and 17.31\% on average F1 since our \textsl{MCNS} is better than \textsl{Shapelets} at capturing subsequences' affection of classification results.

\textbf{\textsl{CausalPrune} vs. No Prune.} The pruned size of the train set is shown in Table 1. We observe under different scales dataset settings by comparing \textsl{CausalPrune} with non-\textsl{CausalPrune}. Datasets are cropped in different proportions, which is related to the size of the dataset.
Furthermore, after the prune, the Acc and F1 on the \textsl{LSTM} have increased about 13-15\% on average Acc and F1, which illustrates that our \textsl{CausalPrune} method can discard harmful and redundant data.

\textbf{\textsl{MCNS} as Presentations.} Additionally, our \textsl{MCNS} can represent time series datasets or specific time series. As shown in Figure 4, one significant use of \textsl{MCNS} is to replace standard folder icons with \textsl{MCNS} graphs that show critical data and relations reflecting the dataset’s content. For labeled time series datasets, we can see why different time series are categorized into different classes and essential features. For unlabeled time series datasets, we can see representative subsequences and causality among them, allowing an analyst to spot patterns and anomalies at a glance. Furthermore, by discarding some factors that do not exist in the specific time series, we have similar representations as in Figure 5.

\begin{figure}[ht]   
	
	\centering
	
	\includegraphics[width=0.92\linewidth,scale=0.70]{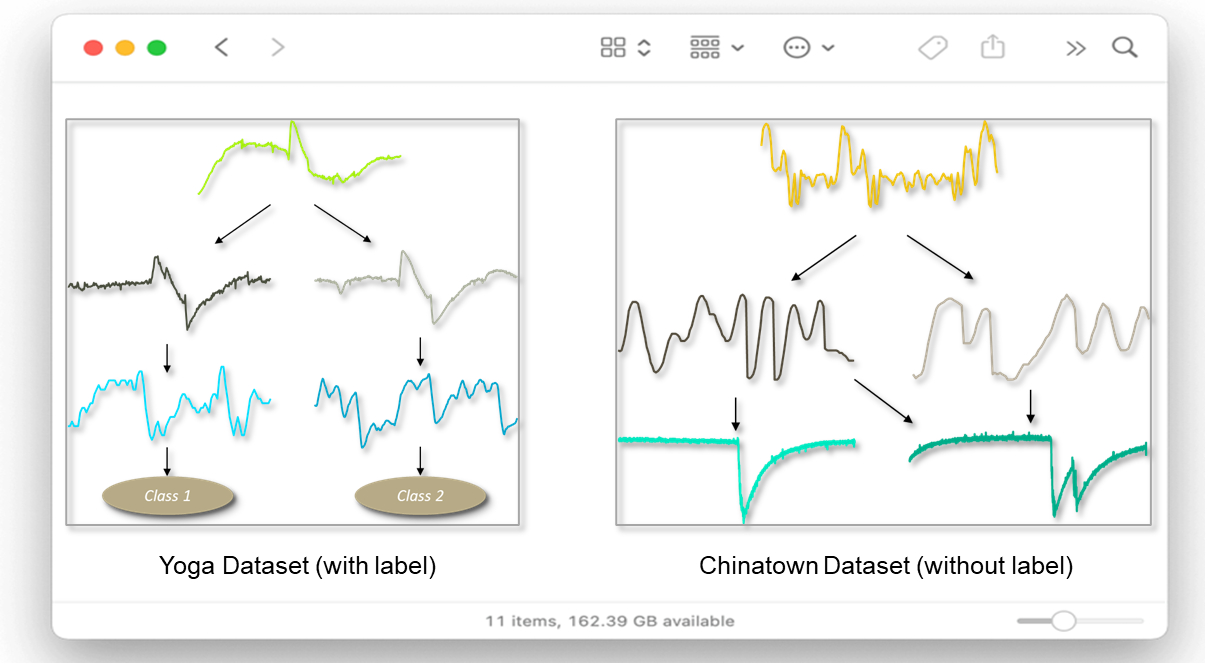}
	
	
	\caption{\textsl{MCNS} representation of labeled time series datasets (left) and unlabelled time series datasets (right), which allows researchers to discover the features and relationships of datasets at a glance.}
	
	\label{intro}
	
\end{figure}
\begin{figure}[h]   
	
	\centering
	
	\includegraphics[width=\linewidth,scale=0.70]{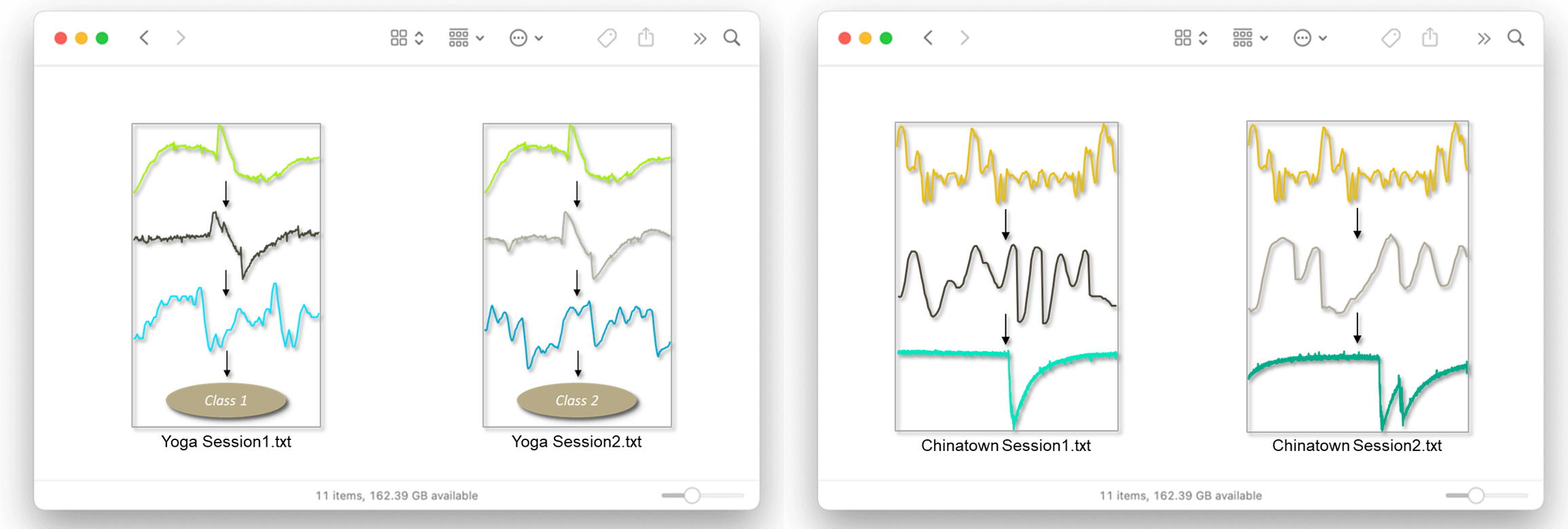}
	
	
	\caption{Two examples of specific labeled (left) and unlabeled (right) data represent the simple representation of our causal natural structures from \textsl{MCNS} on each time series.}
	
	\label{intro}
	
\end{figure}

\section{Influence of Parameters}
We evaluate the impact of parameter settings on \textsl{MCNS}. Specifically, parameters $l$ and $k$ represent the length and number of snippets. Without cherry-picking, we randomly chose two datasets such as FordA and Strawberry, and explored the effectiveness of our approach. All results are averaged over five runs.

We propose a novel metric $CIR$ to evaluate the above parameters, which represents the causal information ratio as follows:
\begin{equation}
    CIR = \frac{\tau}{n},\tau \in n
\end{equation}

where $n$ is the number of clusters, and $\tau$ is the number of causal factors pointing to the classification label. The $CIR$ stands for the ability of a causal graph to represent the original data or dataset.

\begin{figure}[t]
\begin{minipage}[t]{0.49\linewidth}
    \includegraphics[width=\linewidth]{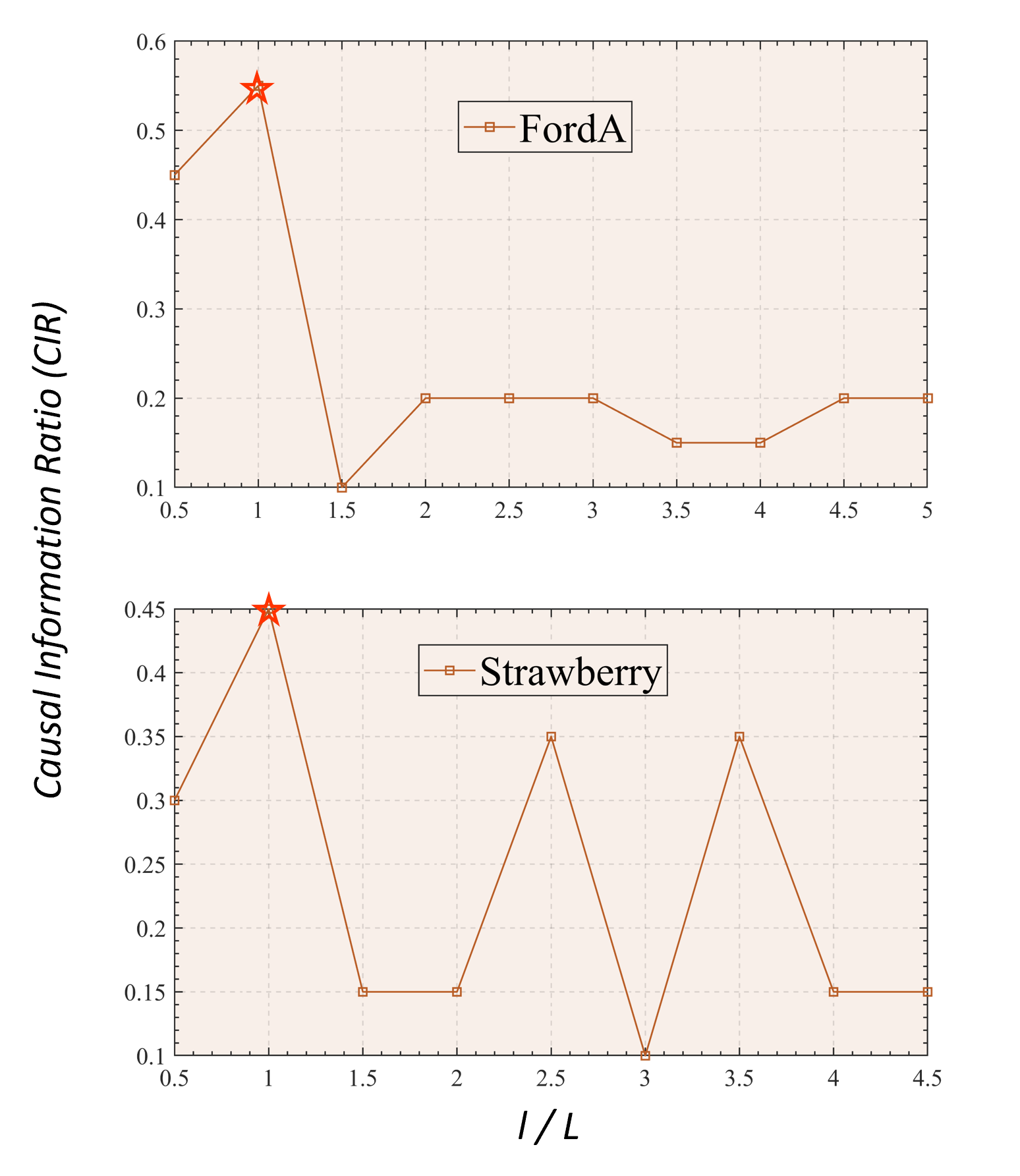} 
    \caption{Influence of $l$: length} 
    \label{f1}
\end{minipage}%
    \hfill%
\begin{minipage}[t]{0.49\linewidth}
    \includegraphics[width=\linewidth]{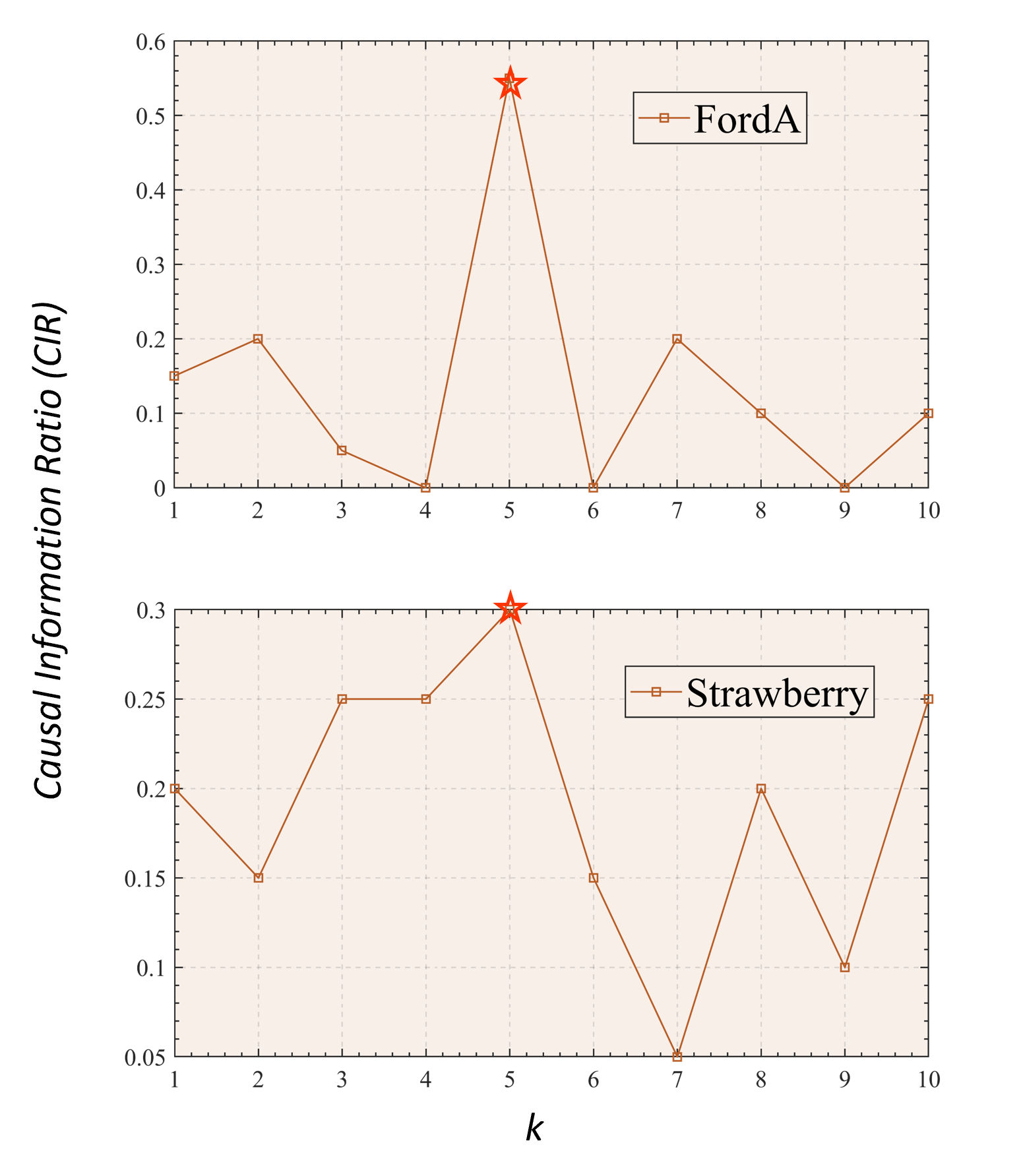}
    \caption{Influence of $k$: number}
    \label{f2}
\end{minipage} 
\end{figure}

\subsection{The Length of Snippets}
Recall that we set $l$ using an FFT-based method. Let the value recommended by this method be $L$. We set $l$ to $0.5 × L$ : $0.5 × L$ : $5 × L$ with all other parameters fixed. 
Then, we evaluate the ratio $CIR$. Figure 6 shows the $CIR$ with varying $l$, where the asterisk is the recommended length. As is shown, when $l$ increases beyond $L$, the causal information ratio gradually decreases and fluctuates, with the best accuracy obtained when $l = L$. Additionally, in $Strawberry$, we can not find snippets with $5L$ length because it is too large. The above validates the effectiveness of our FFT-based approach.

\subsection{The Number of Snippets}
Recall that we set an uncertain number of $k$ for snippets discovery.
For a given time series length $n$, we varied the value $k$ to test the effectiveness of \textsl{MCNS} against $k$. As is shown in Figure 7, our method is relatively sensitive to $k$, where the asterisk is the recommended number. That is because the specific value of the parameter $k$ affects the amount of information. Hence, setting an appropriate $k$ is essential. Inspired by our empirical results, for most real-world datasets, like the two we choose at random, it is proper to set $k$ to around 5.

\section{Conclusion}
Mining causal natural structures inside time series is a challenging problem. To find out the causal natural structures inside time series data, We propose a novel framework called \textsl{MCNS}. It benefits neural networks by refining attention, shape causal selection based classification, and dataset pruning. Extensive experimental results on six real-world datasets from various domains and scales have demonstrated the feasibility and generalization of our approach. The future work will apply \textsl{MCNS} to multidimensional time series and integrate \textsl{MCNS} into diverse NNs. Furthermore, our \textsl{MCNS} can naturally benefit other fields, such as reinforcement learning, adversarial attack, etc.

\bibliographystyle{named}
\bibliography{ijcai23}

\end{document}